\documentclass{article}

\usepackage[nonatbib,final]{neurips_2019_ml4ps}
\usepackage[caption=false,font=footnotesize]{subfig}
\usepackage{float}
\usepackage{stfloats}
\usepackage[export]{adjustbox}
\usepackage{wrapfig}
\usepackage{amsfonts}
\usepackage{amsmath}
\usepackage{changes}
\definechangesauthor[name={Sami}, color=red]{sk}
\usepackage{braket}
\usepackage{bm} 
\newcommand{\vect}[1]{\boldsymbol{\mathbf{#1}}}
\newcommand{\HsubM}{H_{\!M}}
\newcommand{\HsubC}{H_{\!C}}
\DeclareMathOperator*{\argmax}{arg\,max}

\usepackage{multicol}
\usepackage{comment}

\usepackage[utf8]{inputenc} 
\usepackage[T1]{fontenc}    
\usepackage{hyperref}       
\usepackage{url}            
\usepackage{booktabs}       
\usepackage{amsfonts,amsmath}       
\usepackage{nicefrac}       
\usepackage{microtype}      
\usepackage{graphicx}

\usepackage{color}

\title{Reinforcement-Learning-Based Variational Quantum Circuits Optimization for Combinatorial Problems}

\author{Sami Khairy \\
Illinois Institute of Technology\\
skhairy@hawk.iit.edu
\And Ruslan Shaydulin\\
Clemson University\\
 rshaydu@g.clemson.edu\\
 \And Lukasz Cincio \\
  Los Alamos National Laboratory \\
  lcincio@lanl.gov \\
 \And Yuri Alexeev \\
 Argonne National Laboratory \\
 yuri@alcf.anl.gov
 \And Prasanna Balaprakash\\
  Argonne National Laboratory \\
 pbalapra@anl.gov}

\begin{document}

\maketitle

\begin{abstract}
Quantum computing exploits basic quantum phenomena such as state superposition and entanglement to perform computations. The Quantum Approximate Optimization Algorithm (QAOA) is arguably one of the leading quantum algorithms
that can outperform classical state-of-the-art methods in the near term. QAOA is a hybrid quantum-classical algorithm that combines a parameterized quantum state evolution with a classical optimization routine to approximately solve combinatorial problems. The quality of the solution obtained by QAOA within a fixed budget of calls to the quantum computer depends on the performance of the classical optimization routine used to optimize the variational parameters. In this work, we propose an approach based on  reinforcement learning (RL)  to train a policy network that can be used to quickly find high-quality variational parameters for unseen combinatorial problem instances. The RL agent is trained on small problem instances which can be simulated on a classical computer, yet the learned RL policy is generalizable and can be used to efficiently solve larger instances. Extensive simulations using the IBM Qiskit Aer quantum circuit simulator demonstrate that our trained RL policy can reduce the optimality gap by a factor up to $8.61$ compared with other   off-the-shelf optimizers tested.
\end{abstract}

\section{Introduction}

Currently available Noisy Intermediate-Scale Quantum (NISQ) computers have limited error-correction mechanisms and operate on a small number of quantum bits (qubits). Leveraging NISQ devices to demonstrate quantum advantage in the near term requires quantum algorithms that can run using low-depth quantum circuits. The Quantum Approximate Optimization Algorithm (QAOA) \cite{farhi2014quantum}, which has been recently proposed for approximately solving combinatorial problems, is considered one of the candidate quantum algorithms that can outperform classical state-of-the-art methods in the NISQ era \cite{streif2019comparison}. QAOA combines a parameterized quantum state evolution on a NISQ device with a classical optimization routine to find optimal parameters. Achieving practical quantum advantage using QAOA is therefore contingent on the performance of the classical optimization routine. 

QAOA encodes the solution to a classical unconstrained binary assignment combinatorial problem in the spectrum of a cost Hamiltonian $H_C$ by mapping classical binary variables $s_i\in \{-1,1\}$ onto the eigenvalues of the quantum Pauli-Z operator $\hat{\sigma}^z$.  The optimal solution to the original combinatorial problem can therefore be found by preparing the highest energy eigenstate of $H_C$. To this end, QAOA constructs a variational quantum state $\ket{\psi(\vect{\beta},\vect{\gamma})}$ by evolving a uniform superposition quantum state $|\psi \rangle = |+\rangle^{\otimes n}$ using a series of alternating operators $e^{-i\beta_k H_M}$ and $e^{-i\gamma_k H_C}$, $\forall k \in [p]$, 
\begin{equation}
    \ket{\psi{(\vect{\beta},\vect{\gamma})}} =  e^{-i\beta_p \HsubM}e^{-i\gamma_p \HsubC}\cdots e^{-i\beta_1 \HsubM}e^{-i\gamma_1 \HsubC}\ket{+}^{\otimes n},
\label{eq:ansatz}
\end{equation}
\noindent where $\vect{\beta},\vect{\gamma} \in [-\pi, \pi]$ are $2p$ variational parameters, $n$ is the number of qubits or binary variables, and $H_M$ is the transverse field mixer Hamiltonian $H_M = \sum_i \hat{\sigma}_i^x$. 
In order to prepare the highest energy eigenstate of $H_C$, a classical optimizer is used to maximize the expected energy of $H_C$, 
\begin{equation}
    f(\vect{\beta},\vect{\gamma}) = \bra{\psi{(\vect{\beta},\vect{\gamma})}}\HsubC\ket{\psi{(\vect{\beta},\vect{\gamma})}}.
    \label{eq:obj}
\end{equation}
For $p\rightarrow\infty$, $\exists \vect{\beta}_*,\vect{\gamma}_* = \argmax_{\vect{\beta}, \vect{\gamma}}f(\vect{\beta},\vect{\gamma})$ such that the resulting quantum state $\ket{\psi{(\vect{\beta}_*,\vect{\gamma}_*)}}$ encodes the optimal solution to the classical combinatorial problem \cite{farhi2014quantum}. 
QAOA has been applied to a variety of problems, including network community detection~\cite{shaydulin2018network,shaydulin2018community}, portfolio optimization \cite{barkoutsos2019improving}, and graph maximum cut (Max-Cut)~\cite{crooks2018performance,zhou2018quantum}. In this work, we choose graph Max-Cut as a target problem for QAOA because of its equivalence to quadratic unconstrained binary optimization. 

Consider a graph $G=(V,E)$, where $V$ is the set of vertices and $E$ is the set of edges. The goal of Max-Cut is to partition the set of vertices $V$ into two disjoint subsets such that the total weight of edges separating the two subsets is maximized:
\begin{equation} \label{maxcut2}
    \max_{\vect{s}}\sum_{i,j\in V} w_{ij} s_i s_j + c, \qquad s_k\in \{-1,1\}, \forall k ,
\end{equation}
\noindent where $s_k$ is a binary variable that denotes partition assignment of vertex $k$, $\forall k \in [n]$, $w_{ij} = 1$ if $(i,j)\in E$, and $0$ otherwise, and $c$ is a constant. In order to encode \eqref{maxcut2} in a cost Hamiltonian, binary variables $s_k$ are mapped onto the eigenvalues of the Pauli-Z operator $\hat{\sigma}^z$,
\begin{equation} \label{ham}
    H_C = \sum_{i,j\in V} w_{ij} \hat{\sigma}^z_i \hat{\sigma}^z_j.
\end{equation}

The works of \cite{osti_1492737,PhysRevA.97.022304,crooks2018performance} show that QAOA for Max-Cut can achieve approximation ratios exceeding those achieved  by the classical Goemans-Williamson algorithm~\cite{goemans1995improved}. However, QAOA parameter optimization is known to be a hard problem because \eqref{eq:obj} is nonconvex with low-quality nondegenerate local optima for high $p$ \cite{shaydulin2019multistart,zhou2018quantum}. Existing works explore many approaches to QAOA parameter optimization, including a variety of off-the-shelf gradient-based~\cite{romero2018strategies,zhou2018quantum,crooks2018performance} and derivative-free methods~\cite{wecker2016training,yang2017optimizing,shaydulin2019multistart}. Noting that the optimization objective \eqref{eq:obj} is specific to a given combinatorial instance through its cost Hamiltonian \eqref{ham}, researchers have approached the task of finding optimal QAOA parameters as an instance-specific task. To the best of our knowledge, approaching QAOA parameter optimization as a learning task is underexplored, with  few recent works \cite{verdon2019learning}.

Thus motivated, in this work we propose a method based on reinforcement learning (RL) to train a policy network that can learn to exploit geometrical regularities in the QAOA optimization objective, in order to efficiently optimize new QAOA circuits of unseen test instances. The RL agent is trained on a small Max-Cut combinatorial instance that can be simulated on a classical computer, yet the learned RL policy is generalizable and can be used to efficiently solve larger instances from different classes and  distributions. By conducting extensive simulations using the IBM Qiskit Aer quantum circuit simulator, we show that our trained RL policy can reduce the optimality gap by a factor of up to $8.61$ compared with  commonly used off-the-shelf optimizers.

\section{Proposed approach}

Learning an optimizer to train machine learning models has recently attracted considerable research interest. 
The motivation is to design optimization algorithms that can exploit structure within a class of problems, which is otherwise unexploited by hand-engineered off-the-shelf optimizers. In existing works, the learned optimizer 
is implemented by a long short-term memory network \cite{andrychowicz2016learning} or a policy network of an RL agent \cite{li2016learning}. Our proposed approach to QAOA optimizer learning departs from that of \cite{li2016learning} in the design of the reward function and the policy search mechanism. 

In the RL framework, an autonomous agent learns how to map its state $s \in \mathcal{S}$, to an action $a \in \mathcal{A}$, by repeated interaction with an environment. The environment provides the agent with a reward signal $r \in \mathbb{R}$, in response to its action.
A solution to the RL task is a stationary Markov policy that maps the agent's states to actions, $\pi(a|s)$, such that the expected total discounted reward is maximized \cite{sutton2018reinforcement}. Learning a QAOA optimizer can therefore be regarded as learning a policy that produces iterative QAOA parameter updates, 
based on the following state-action-reward formulation, 
\begin{enumerate}
\item  $\forall s_t \in \mathcal{S}$, $s_t = \{\Delta f_{tl}, \Delta \vect{\beta}_{tl},\Delta \vect{\gamma}_{tl} \}_{l=t-1,...,t-L}$; in other words, the state space is the set of finite differences in the QAOA objective and the variational parameters between the current iteration and $L$ history iterations, $\mathcal{S} \subset \mathbb{R}^{(2p+1)L}$.
\item $\forall a_t \in \mathcal{A}$, $a_t = \{\Delta \vect{\beta}_{tl}, \Delta \vect{\gamma}_{tl} \}_{l=t-1}$; in other words, the action space is the set of step vectors  used to update the variational parameters, $\mathcal{A} \subset \mathbb{R}^{2p}$.
\item $\mathcal{R}(s_t, a_t, s_{t+1}) = f(\vect{\beta}_t + \Delta \vect{\beta}_{t,tl}, \vect{\gamma}_t + \Delta \vect{\gamma}_{tl}) - f(\vect{\beta}_t, \vect{\gamma}_t), l=t-1$; in other words, the reward is the change in the QAOA objective between two consecutive iterations. 
\end{enumerate}

The motivation for our state space formulation comes from the fact that parameter updates at $\vect{x}_t = (\vect{\beta}_{t},\vect{\gamma}_{t})$ should be in the direction of the gradient at $\vect{x}_t$ and the step size  should be proportional to the Hessian at $\vect{x}_t$, both of which can be numerically approximated by using the method of finite differences. The RL agent's role is then to find an optimal reward-maximizing mapping to produce the step vector $a_t = \{\Delta \vect{\beta}_{tl}, \Delta \vect{\gamma}_{tl} \}_{l=t-1}$, given some collection of historical differences in the objective and parameters space, $\{\Delta f_{tl}, \Delta \vect{\beta}_{tl},\Delta \vect{\gamma}_{tl} \}_{l=t-1,...,t-L}$. Note that the cumulative rewards are maximized when $\mathcal{R}(s_t, a_t, s_{t+1}) \geq 0$, which means the QAOA objective has been increased between any two consecutive iterates. The reward function adheres to the Markovian assumption and encourages the agent to produce parameter updates that yield higher increase in the QAOA objective \eqref{eq:obj}, if possible, while maintaining conditional independence on historical states and actions.
\section{Experiments}
\begin{wrapfigure}{r}{0.3\textwidth} \label{fig:GRAPHS}
  \begin{center}
    \includegraphics[width=0.3\textwidth]{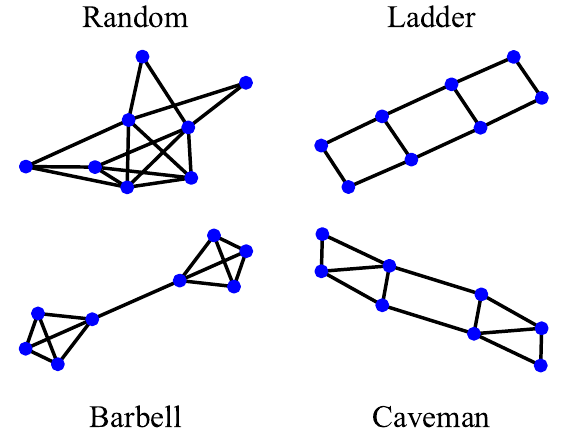}
  \end{center}
  \caption{Sample graph instances, TBLR:  $G_R(n_R=8, e_p=0.5)$, $ G_L(n_L=4)$, $G_B(n_B=4)$, $G_C(n_C=2, n_k=4)$.} 
\end{wrapfigure}
We train our proposed approach on a small $n_R=8$ qubit graph instance drawn from an Erdos Renyi random graph with edge generation probability $e_p=0.5$. Training was performed over 750 epochs, where each
epoch corresponds to $8,192$ QAOA circuit simulations  executed by using IBM Qiskit Aer simulator \cite{Qiskitshort}. Within each epoch are $1,28$ episodes. Each training episode corresponds to a trajectory of length $T=64$ that is sampled from a depth-$p$ QAOA objective \eqref{eq:obj} for the training instance. At the end of each episode, the trajectory is cut off and is randomly restarted.

For policy search, we adopt the actor-critic Proximal Policy Optimization (PPO) algorithm \cite{schulman2017proximal}, which uses a clipped surrogate advantage objective as a training objective. 
To further mitigate policy updates that can cause policy collapse, we adopt a simple early stopping method, which terminates gradient optimization on the PPO objective when the mean KL-divergence between the new and old policy hits a predefined threshold. Fully connected multilayer perceptron networks with two $64$-neuron hidden layers for both the actor and critic networks are used. 
A Gaussian policy with a constant noise variance of $e^{-6}$ is adopted throughout training. At testing, the trained policy network corresponding to the mean of the learned Gaussian policy is used, without noise.

To test the performance of the learned RL policy, we chose a large set $G_\text{Test}$ of Max-Cut test instances, coming from different sizes, classes, and distributions. Specifically, four classes of graphs are considered: (1) Erdos Renyi random graphs $G_R(n_R, e_p)$, $n_R \in \{8,12,16,20 \}$, $e_p \in \{0.5,0.6,0.7,0.8\}$, seed $=\{1,2,3,4\}$; (2) ladder graphs $G_L(n_L)$, where $n_L \in \{2,3,4,5,6,7,8,9,10,11 \}$ is the length of the ladder; (3) barbell graphs $G_B(n_B)$, formed by connecting two complete graphs $K_{n_B}$ by an edge, where $n_B \in \{3,4,5,6,7,8,9,10,11\}$; and (4) caveman graphs $G_C(n_C, n_k)$, where $n_C$ is the number of cliques and $n_k$ is the size of each clique, $\{(n_C,4):n_C \in \{3,4,5\}\}$, $\{(n_C,3):n_C \in \{3,5,7\}\}$,$\{ (2,n_K): n_K \in \{3,4,5,6,7,8,9,10\} \}$. Thus, $|G_\text{Test}|=97$. Figure 1 shows sample graph instances in $G_\text{Test}$. $G_\text{Test}$ is chosen to demonstrate that combining our proposed RL-based approach with QAOA can be a powerful tool for amortizing the QAOA optimization cost across graph instances, as well as demonstrating the generalizability of the learned policy.

\section{Results}
\begin{figure}[ht!]
\hfill
\subfloat[$G_R(16,0.5)$]{\includegraphics[width=0.25\columnwidth]{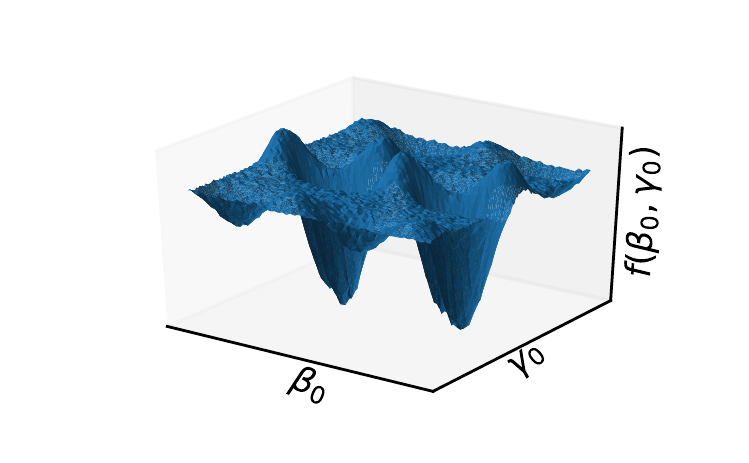}}
\hfill
\subfloat[$G_L(8)$]{\includegraphics[width=0.25\columnwidth]{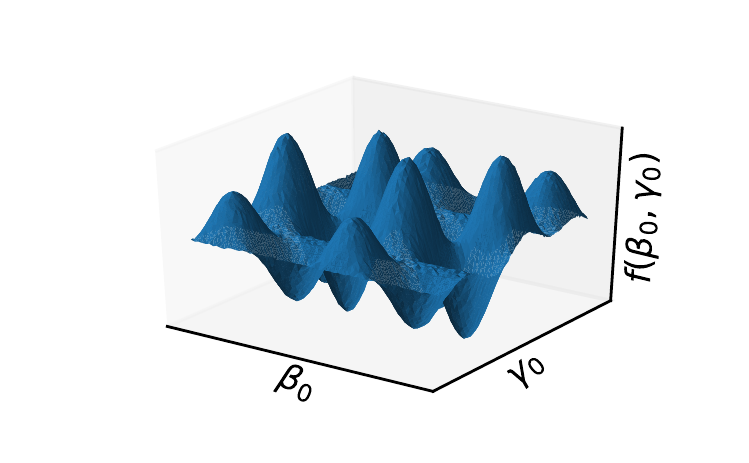}} 
\hfill
\subfloat[$G_B(8)$]{\includegraphics[width=0.25\columnwidth]{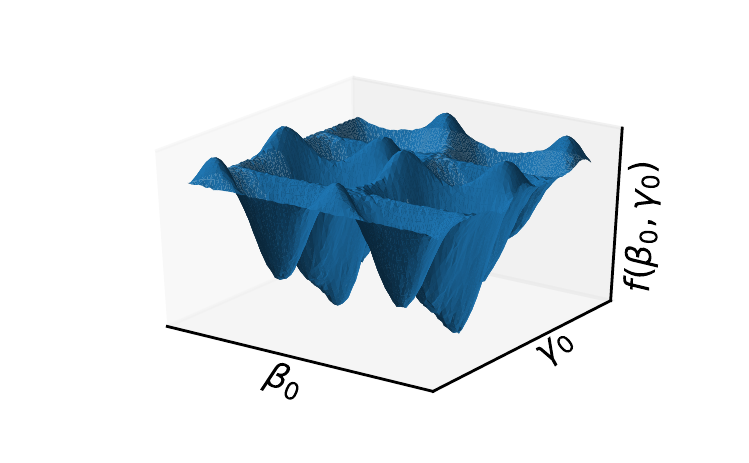}}
\hfill
\subfloat[$G_C(2,8)$]{\includegraphics[width=0.25\columnwidth]{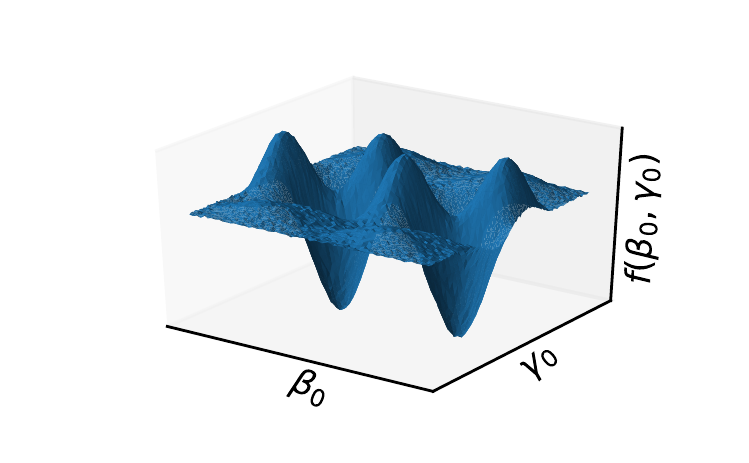}} 
\hfill
\caption{QAOA energy landscapes for $p=1$.}
\label{fig:energyLandscapes}
\end{figure}
\begin{figure}[ht]
\centering
\hfill
\subfloat[Objective value]{\includegraphics[width=0.5\columnwidth]{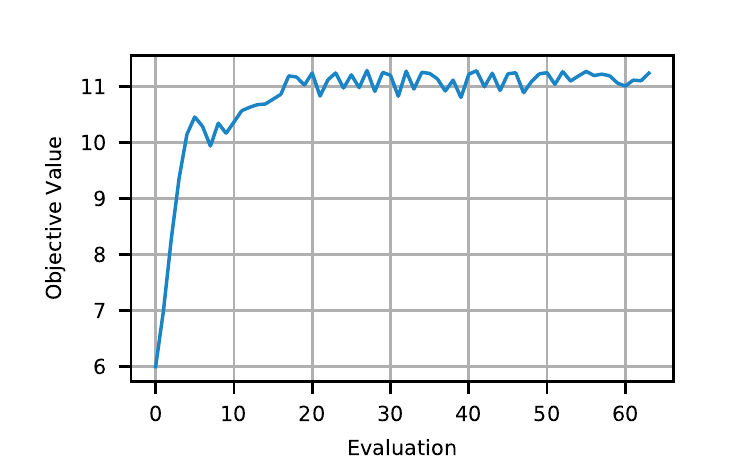}}
\centering
\hfill
\subfloat[Trajectory]{\includegraphics[width=0.5\columnwidth]{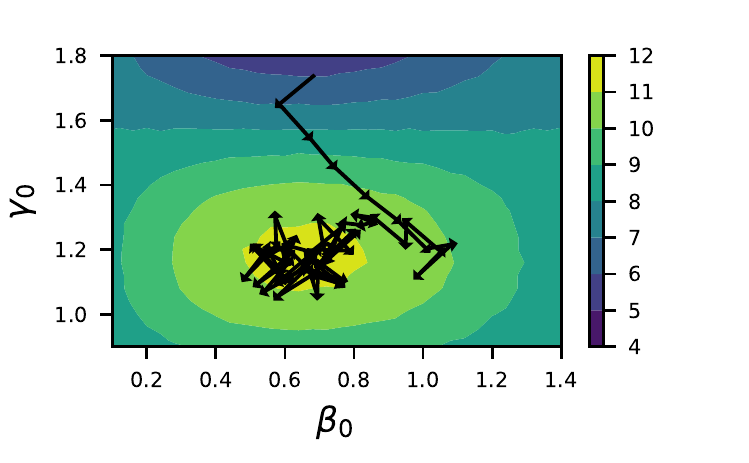}} 
\centering
\hfill
\subfloat[QAOA Approximation Ratio]{\includegraphics[width=0.5\columnwidth]{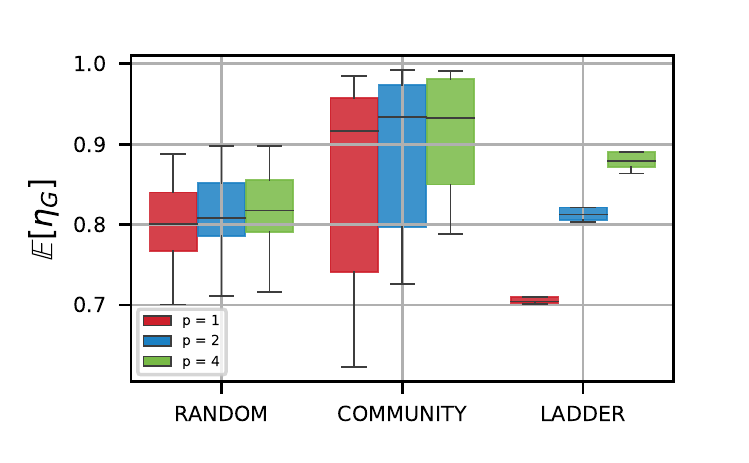}}
\caption{Visualization of the trajectory produced by the learned RL policy on one of the test instances for the QAOA circuit of $p=1$ (a)--(b), and the approximation ratio performance of QAOA with respect to classical optimal on graph instances in $G_\text{Test}$ (c).}
\label{fig:trajVis}
\end{figure}
\begin{figure}[hb]
\centering
\hfill
\subfloat[$p=1$]{\includegraphics[width=0.50\columnwidth]{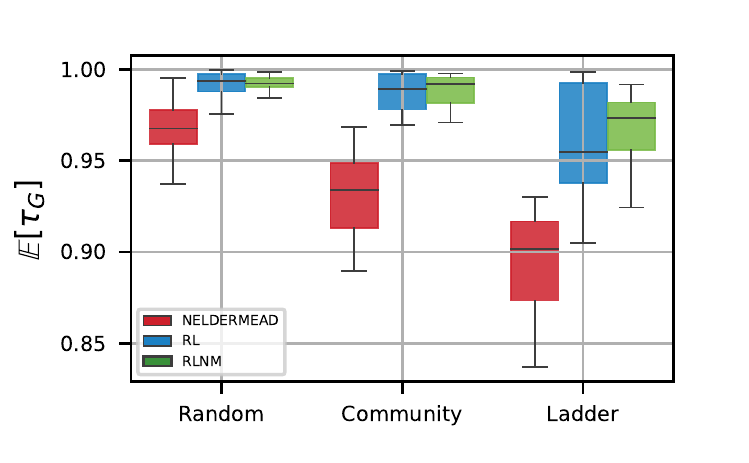}}
\centering
\hfill
\subfloat[$p=2$]{\includegraphics[width=0.50\columnwidth]{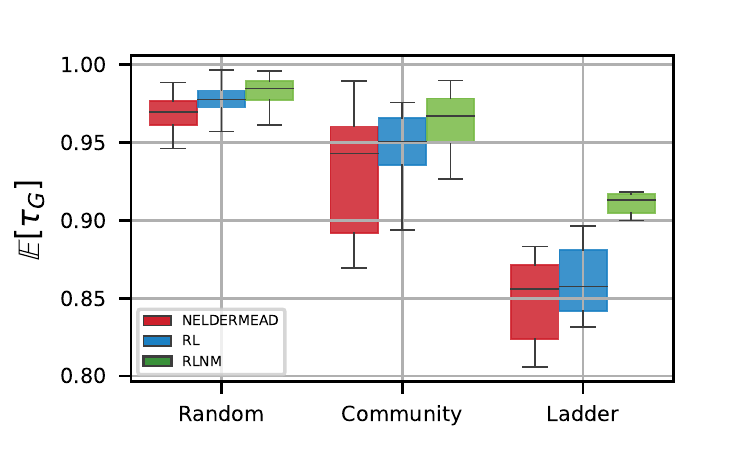}} 
\centering
\hfill
\subfloat[$p=4$]{\includegraphics[width=0.50\columnwidth]{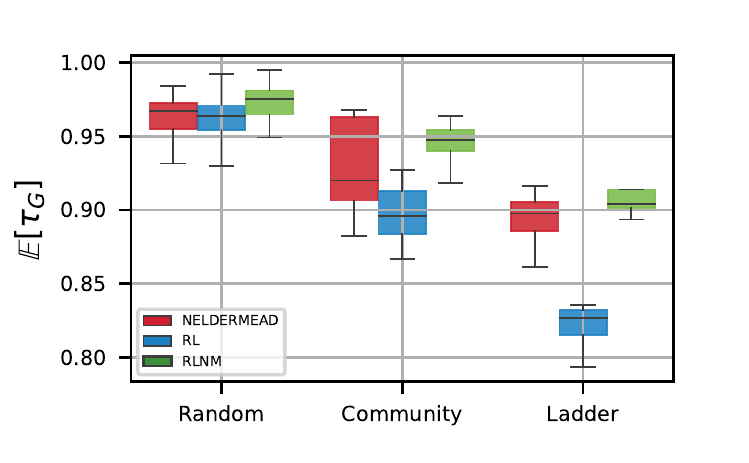}}
\caption{Expected optimality ratio performance of Nelder-Mead, learned RL optimization policy, and the combined RLNM for a given QAOA circuit depth $p \in \{1,2,4\}$ on graph instances in $G_\text{Test}$.}
\label{fig:optimizerPerformance}
\end{figure}

In Figure \ref{fig:energyLandscapes}, the expected energy in \eqref{eq:obj} for a depth $(p=1)$ QAOA circuit is shown for some graph instances. We observe that the expected energy is nonconvex in the parameter space. Figures \ref{fig:trajVis}(a) and (b) visualize the trajectory produced by the learned RL policy on one of the test instances. We can see that the learned policy produces trajectories that quickly head to the maximum (in about $20$ iterations in this example), yet a wiggly behavior is observed afterwards. Figure \ref{fig:trajVis}(c) shows a boxplot of the expected approximation ratio performance, $\mathbb{E}[\eta_G] = \mathbb{E}[f/C_\text{opt}]$, of QAOA with respect to the classical optimal $C_\text{opt}$ found by using brute-force methods across different graph instances in $G_\text{Test}$, which are grouped in three subgroups: (1) random graphs, which contains all graphs of the form $G_R(n_R,e_p)$; (2) community graphs, which contains graphs of the form $G_C(n_C, n_k)$ and $G_B(n_B)$; and (3) ladder graphs, which contains graphs of the form $G_L(n_L)$. We can see that increasing the depth of the QAOA circuit improves the attained approximation ratio.
Next, we benchmark the performance of our trained RL-based QAOA optimization policy (referred to as RL) by comparing its performance with that of a commonly used derivative-free off-the-shelf optimizer, namely, Nelder-Mead~\cite{nelder1965simplex}, on graph instances in $G_\text{Test}$. Starting from 10 randomly chosen variational parameters in the domain of \eqref{eq:obj}, each optimizer is given $10$ attempts with a budget of $B=192$ quantum circuit evaluations. In addition, we use the learned RL policy to generate trajectories of length $B/2$, and we resume the trajectory from the best parameters found using Nelder-Mead for the rest of $B/2$ evaluations (referred to as RLNM). This approach is motivated by the observed behavior of the learned RL policy in \ref{fig:trajVis}(b). In Figures \ref{fig:optimizerPerformance} (a)--(c), we present a boxplot of the expected optimality ratio, $\mathbb{E}[\tau_{G}] =  \mathbb{E}[f/f_\text{opt}]$, where the expectation is with respect to the highest objective value attained by a given optimizer in each of its $10$ attempts. The optimal solution to a graph instance in $G_\text{Test}$ is the largest known $f$ value found by any optimizer in any of its $10$ attempts for a given depth $p$. We can see that the median optimality ratios achieved by RL and RLNM outperform that of Nelder-Mead for $p=\{1,2\}$ and $p=\{1,2,4\}$, respectively. The median optimality gap reduction factor of RLNM with respect to Nelder-Mead 
ranges from $1.16$ to $8.61$ depending on the graph subgroup and QAOA circuit depth $p$. 

\section{Conclusion}
\label{sec:con}

In this paper, we addressed the problem of finding optimal QAOA parameters as a learning task. We propose an RL-based approach that can learn a policy network that exploits regularities in the geometry of QAOA instances to efficiently optimize new QAOA circuits. We demonstrate that there is a learnable policy that generalize well across different instances of different sizes, even when the agent is trained on small instances. The learned policy can reduce the optimality gap by a factor up to $8.61$ compared with other  off-the-shelf optimizers tested.

\subsubsection*{Acknowledgments}
This material is based upon work supported by the U.S. Department of Energy (DOE), Office of Science, Office of Advanced Scientific Computing Research, under Contract DE-AC02-06CH11357. 
This research was funded in part by and used resources of the Argonne Leadership Computing Facility, which is a DOE Office of Science User Facility supported under Contract DE-AC02-06CH11357.
We gratefully acknowledge the computing resources provided on Bebop, a high-performance computing cluster operated by the Laboratory Computing Resource Center at Argonne National Laboratory.

\onecolumn{
\bibliographystyle{unsrt}
\bibliography{qaoa}}

\begin{center}
    \framebox{\parbox{5in}{
    The submitted manuscript has been created by UChicago Argonne, LLC, Operator of Argonne National Laboratory (``Argonne''). Argonne, a U.S. Department of Energy Office of Science laboratory, is operated under Contract No. DE-AC02-06CH11357. The U.S. Government retains for itself, and others acting on its behalf, a paid-up nonexclusive, irrevocable worldwide license in said article to reproduce, prepare derivative works, distribute copies to the public, and perform publicly and display publicly, by or on behalf of the Government. The Department of Energy will provide public access to these results of federally sponsored research in accordance with the DOE Public Access Plan. \url{http://energy.gov/downloads/doe-public-access-plan}}}
    \normalsize
\end{center}

\end{document}